\documentclass[11pt]{article}

\usepackage[preprint]{acl}

\usepackage{times}
\usepackage{latexsym}
\usepackage{amsmath,amssymb}
\usepackage[T1]{fontenc}
\usepackage[utf8]{inputenc}
\usepackage{microtype}
\usepackage{inconsolata}
\usepackage{graphicx}
\usepackage{mathptmx}
\usepackage{booktabs}
\usepackage{multirow,xcolor}
\usepackage{float}

\definecolor{resultgreen}{RGB}{0,128,0}
\newcommand{\accgain}[2]{%
  #1\,{\scriptsize\textcolor{resultgreen}{$\uparrow\,#2$}}%
}

\title{Task Vectors Can Be Linearly Decomposed into Interpretable and Controllable Components}

\author{
  Yukiya Horiba\textsuperscript{1} \quad
  Hakaze Cho \textsuperscript{2, 3} \quad
  Takahiro Yakoh \textsuperscript{1} \\
  \textsuperscript{1}Keio University
  \textsuperscript{2}RIKEN
  \textsuperscript{3}Tohoku University \\
}

\begin{document}
\maketitle

\begin{abstract}
\textbf{T}ask \textbf{V}ectors (TVs) provide a compact way to steer large language models, but they are typically treated as atomized directions merely extracted from hidden states, leaving their internal functional structure poorly understood.
Motivated by prior work that decomposes in-context learning into \textbf{T}ask \textbf{R}ecognition (TR) and \textbf{T}ask \textbf{L}earning (TL), we ask whether a single TV can likewise be decomposed into functionally meaningful components.
We formulate a TR/TL-based decomposition from candidate-label unembeddings, separating each TV into TR, TL, and residual components, and try to optimize the TR and TL coefficients using Bayesian optimization for more effective intervention.
We evaluate this formulation across multiple model families, classification tasks, and both instruction- and demonstration-based TVs.
Across these settings, TR/TL-based optimization identifies intervention directions that outperform the original TVs.
Although optimization over random directions can also improve performance, TR/TL-based search yields more reliable gains, particularly for instruction-based TVs.
Moreover, isolated interventions on the decomposed components produce distinct logit-level effects that are consistent with their intended functional roles.
Together, these results suggest that TVs are not monolithic steering directions, but can be decomposed into functionally meaningful components that support both interpretation and control.
\end{abstract}


\begin{figure*}[!t]
    \centering
    \includegraphics[width=0.96\textwidth]{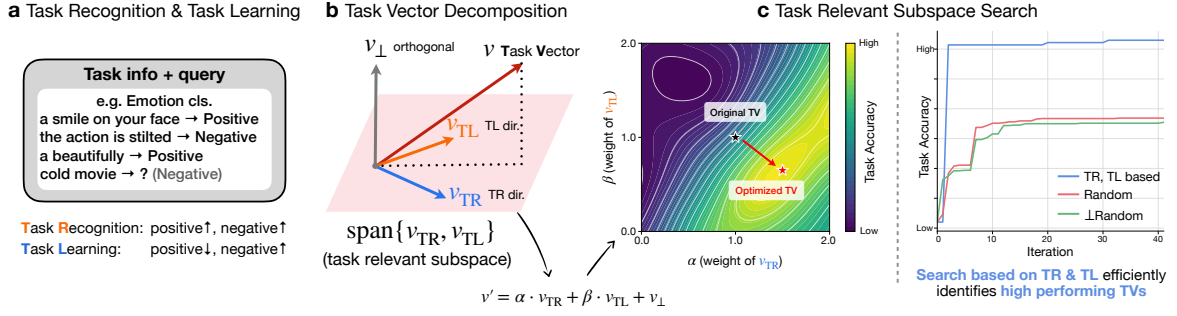}
    \vspace{-0.5\baselineskip}
    \caption{
        \textbf{Overview of our approach}.
        (a) TR and TL capture complementary aspects of in-context learning. In a narrow sense, TR processing finds the label candidates like ``Positive'' and ``Negative'', and TL find which label should be output in the candidates.
        (b) We decompose a TV into TR, TL, and residual components.
        (c) Optimizing the TR/TL coefficients improves task performance.
    }
    \label{fig:teaser}
\end{figure*}

\section{Introduction}
\label{sec:introduction}

\textbf{T}ask \textbf{V}ectors (TVs) provide a compact representation of task information that can be reused to steer model behavior without updating model parameters~\cite{hendel2023context}.
Despite their effectiveness, TVs are typically treated as merely atomized blackbox vectors extracted from hidden states, which leaves a basic question open: how is task information functionally organized within a TV? Understanding this structure would make TVs not only more interpretable, but also effective.
Existing work has composed multiple steering directions into a low-dimensional intervention space~\cite{han2026steer2adapt}, but still little is known about whether a single TV itself can be decomposed into functionally meaningful components.

Conceptually, task execution in LLMs, exemplified by \textbf{I}n-\textbf{c}ontext \textbf{L}earning (ICL), is not necessarily functionally monolithic. As shown in Fig.~\ref{fig:teaser}-a, prior work distinguishes \textbf{T}ask \textbf{R}ecognition (TR), which identifies the task and leverages pretrained knowledge, from \textbf{T}ask \textbf{L}earning (TL), which learns input-label mappings from demonstrations~\cite{pan2023context}.
Recent mechanistic analyses further associate these functions with distinct internal computations~\cite{yang2026localizing}.
If these functional roles contribute to the task information encoded by a TV, they may provide a meaningful basis for exposing its internal structure.

We therefore ask whether the conceptual decomposition of ICL into TR and TL can be extended to geometry decomposition of TVs: \textit{can a TV be decomposed into functionally distinct TR- and TL-related components, and can these components provide meaningful control over the TV?}
As illustrated in Fig.~\ref{fig:teaser}-b, we construct TR/TL-oriented axes from candidate-label geometry and decompose each TV into TR, TL, and residual components.
We parameterize intervention directions using the TR and TL magnitudes while keeping the residual fixed, which defines a two-dimensional, functionally structured intervention space anchored by the original TV, with each point in this space represents a distinct weighted combination of the TR and TL components. Additionally, as shown in Fig.~\ref{fig:teaser}-c, in this space, we conduct Bayesian optimization to find TV with better performance than original TV.

We evaluate the performance of the optimized TVs across multiple models and tasks, and find that Bayesian optimized TVs within the TR/TL-based space largely outperform the original TVs across a broad range of settings.
At the same time, optimization over random directions can occasionally yield substantial improvements, showing that high-performing directions are not unique to the TR/TL directions.
But TR/TL-based optimization tends to yield larger and more reliable gains, suggesting that the decomposition provides useful functional structure for guiding the search rather than uniquely determining an optimal direction.
These results suggest that conceptual decomposition of ICL mechanisms can provide an interpretable geometry basis for analyzing and controlling TVs.

Our contributions are threefold.
\textbf{First}, we formulate a decomposition of a single TV into TR, TL, and residual components, linking its structure to the roles of TR and TL with supporting empirical analysis.
\textbf{Second}, we show that this decomposition provides an interpretable parameterization for controlling the TV, in which optimizing the TR and TL coefficients improves task performance across multiple models and tasks.
\textbf{Third}, we draw on these findings to offer theoretical insights into how TR and TL jointly contribute to task execution.


\section{Related Work}

\paragraph{TV in ICL.}
In standard ICL, a model infers a task from examples provided in the prompt, commonly referred to as demonstrations as shown in Fig.~\ref{fig:teaser}-a.
\citet{hendel2023context} has shown that task information induced by these demonstrations can be encoded in intermediate activations and reusable even after the demonstrations are removed.
Subsequent work explored related activation space reprentations: \citet{todd2024function} identified function vectors from causally relevant attention heads, while \citet{Liu2024incontext} constructed in-context vectors for steering model behavior.
More recent studies have shifted toward understanding and improving these representations.
\citet{yang2025unifying} connected TV behavior to hidden state geometry and attention head mechanisms, and \citet{yang2026task} showed that TVs learned from scratch can outperform standards extracted ones.
Together, these studies suggest that task information in activation space is structured and manipulable, while how different aspects of this information are organized within a TV remains only partially understood.

\paragraph{Mechanisms of ICL.}
Prior work has shown that ICL can rely on both pretrained task knowledge and input-label mappings learned from demonstrations.
\citet{min2022rethinking} found that models can often retain substantial ICL performance even when demonstration labels are falsified, highlighting the importance of task-level information beyond the input-label mapping itself.
\citet{wei2023larger} further showed that larger models can learn novel input-label mappings from context even when they conflict with pretrained semantic priors. As shown in Fig.~\ref{fig:teaser}-a, 
\citet{pan2023context} characterized ICL through two complementary abilities: TR, which identifies the task and leverages pretrained knowledge, and TL, which learns new input label mappings from demonstrations. In a narrow sense of classification tasks, TR determines the set of label candidates as a task representation, and TL select one label in the candidates as the defined mapping from the task.
\citet{wang2025investigating} further studied their development during pretraining and found that TR and TL can compete with each other.
\citet{yang2026localizing} localized TR and TL to distinct attention heads and showed that they play complementary roles in shaping hidden state geometry.
These studies suggest functionally distinct ICL process reflected in model computation.

\paragraph{Bayesian Optimization for Hidden State Interventions.}
\textbf{B}ayesian \textbf{O}ptimization (BO) has been used to tune different aspects of inference time interventions without gradient based model updates.
\citet{gao2025shaping} use BO to select layers and intervention parameters for activation based jailbreak defense, while \citet{siddique2026shifting} use it to identify effective data configurations for constructing steering vectors.
More directly, Steer2Adapt~\cite{han2026steer2adapt} constructs a low dimensional semantic subspace from multiple steering vectors and use BO to optimize their composition coefficients.
BOReFT~\cite{agarwal2026boreft} further learns a low dimensional space of hidden state interventions and uses it as the search domain for BO.
These studies use BO to optimize internal interventions, while the structure and interpretation of the search space remain central design choices.


\section{Methodology}
\label{sec:method}

\paragraph{Problem Setup.}
We consider classification tasks, where each task \(t\) has \(K\) candidate labels \(\mathcal{Y}_t\) and each instance consists of an input \(x_i\) and label \(y_i\in\mathcal{Y}_t\).
Given a frozen language model, task information is provided through either an instruction or labeled demonstrations together with the query.
For example, in sentiment classification, an instruction may explicitly ask the model to classify the query as \textit{positive} or \textit{negative}, whereas the demonstration setting provides only labeled examples, such as ``A delightful film.'' \(\rightarrow\) \textit{positive}, without an explicit task instruction.
The baseline without any task information provided is query-only prompt, which contains the same query \(x_i\) but no instruction or demonstrations, and
A TV captures the hidden-state change between the task-conditioned and query-only prompts at Transformer layer \(l\).
Our goal is to decompose this TV into functionally distinct components and use them to obtain more effective intervention directions then the original TVs.

\paragraph{Overview.}
Our methodology has three steps.
First, we extract the original TV from the hidden-state difference between task-conditioned and query-only prompts (Sec.~\ref{sec:tv_construction}).
Second, we derive TR/TL-oriented directions from candidate-label unembedding geometry and decompose the TV into TR, TL, and residual components (Sec.~\ref{sec:decomposition}).
Third, we optimize the TR and TL magnitudes while retaining the residual, using Bayesian optimization to identify higher-performing directions within the resulting structured intervention space (Sec.~\ref{sec:optimization}).

\subsection{TV Extraction}
\label{sec:tv_construction}

We first extract the original TV that serves as the starting point for decomposition and optimization.
Adapting prior TV extraction methods~\cite{hendel2023context, yang2026task}, we compare task-conditioned and query-only hidden states for the same query.
For task \(t\), let \(c\in\{\mathrm{inst},\mathrm{demo}\}\) indicate whether the task information is provided by an instruction or labeled demonstrations.
Given \(N\) queries used for TV extraction, let \(x_i\) denote the \(i\)-th query.
For each \(x_i\), we compare the task-conditioned prompt with the corresponding query-only prompt containing the same query, and define the query-wise TV at layer \(l\) as
\begin{equation}
v_{\mathrm{TV},c,t,i}^{(l)}
=
h_{c,t,i}^{(l)}
-
h_{\emptyset,i}^{(l)},
\label{eq:query_tv}
\end{equation}
where \(h_{c,t,i}^{(l)}\) and \(h_{\emptyset,i}^{(l)}\) denote the final-token hidden states of the task-conditioned and query-only prompts (where the task information is not provided), respectively. Then, we obtain the TV by averaging over the \(N\) queries:
\begin{equation}
v_{\mathrm{TV},c,t}^{(l)}
=
\frac{1}{N}
\sum_{i=1}^{N}
v_{\mathrm{TV},c,t,i}^{(l)}.
\label{eq:tv}
\end{equation}
This TV captures the average hidden-state change induced by task-related context on a same set of the query.

\subsection{TR/TL-Based Decomposition}
\label{sec:decomposition}

A TV provides a single direction that summarizes the representation change induced by task-related context, but does not reveal how different functions are organized within TV.
We therefore decompose the TV into TR, TL, and residual components, so that the contributions associated with TR and TL can be analyzed and controlled separately.

\paragraph{TR/TL-Inspired Label Geometry.}
Prior work distinguishes TR, which identifies the task and its candidate-label space, from TL, which captures the input-label mapping needed to select among candidate labels.
We use this functional distinction as the basis of our decomposition.
Rather than prespecifying task-specific semantic concepts that may help solve a downstream task, we derive the decomposition directly from the candidate-label structure of each task.
Specifically, we represent TR by the structure shared across candidate-label unembedding directions and TL by the contrast between the correct label and the remaining candidate labels.
These directions are intended as task-general, TR/TL-oriented operationalizations rather than complete descriptions of the underlying TR and TL mechanisms.

Let \(W_{\mathrm{label},t}\in\mathbb{R}^{K\times d}\) contain the rows of the model's output projection (unembedding) matrix corresponding to the \(K\) candidate labels of task \(t\), where \(d\) is the hidden-state dimension.
We first define the TR/TL directions from the \(W_{\mathrm{label},t}\in\mathbb{R}^{K\times d}\).
The TR direction assigns the same coefficient to all candidate labels, whereas the TL direction contrasts a target label \(y\) with the other candidates:
\begin{equation}
b_{\mathrm{TR}}
=
\mathbf{1}_{K},
\qquad
b_{\mathrm{TL}}(y)
=
e_y
-
\frac{1}{K-1}
\sum_{j\neq y} e_j,
\label{eq:label_axes}
\end{equation}
where \(\mathbf{1}_{K}\) is the all-ones vector and \(e_y\) is the one-hot vector for label \(y\). We then map these label-space directions into the hidden-state space through the candidate-label unembedding vectors:
\begin{equation}
\begin{aligned}
d_{\mathrm{TR},t}
&=
W_{\mathrm{label},t}^{\top} b_{\mathrm{TR}},
\\
d_{\mathrm{TL},t}(y)
&=
W_{\mathrm{label},t}^{\top} b_{\mathrm{TL}}(y).
\end{aligned}
\label{eq:hidden_axes}
\end{equation}
These directions are defined by the task's candidate labels, not by the query text.
For multiclass tasks, we average the projected TL components in Eq.~\ref{eq:multiclass_tl} to obtain a shared component that remains fixed during subsequent interventions.
These directions have the same dimensionality as TVs at every Transformer layer.
We next specify the projection geometry used to decompose layer-specific TVs along these output-derived directions.

\paragraph{Metric-Aware Projection.}
Because subsequent layers can transform the intervention before it reaches the output, we use the direct candidate-label unembeddings as a shared proxy for defining the projection geometry.
However, the unembedding matrix is located on the last layer of Transformer model, whereas a TV may be extracted at any Transformer layer, which causes potential distributional misalignment.
We therefore need to map the last-layer TR/TL directions to intermediate-layer, to achieve effective decomposition and controlling to TVs from any layer.
Specifically, we weight projection errors according to their local effect on the candidate-label distribution.

To construct this layer-specific geometry, we define a reference distribution by applying the candidate-label unembedding to the original TV:
\begin{equation}
p_{c,t}^{(l)}
=
\operatorname{softmax}
\left(
W_{\mathrm{label},t}
v_{\mathrm{TV},c,t}^{(l)}
\right).
\label{eq:p}
\end{equation}
We then define the hidden-state metric
\begin{equation}
\begin{aligned}
G_{c,t}^{(l)}
&=
W_{\mathrm{label},t}^{\top}
\Bigl[
\operatorname{diag}(p_{c,t}^{(l)})
-
p_{c,t}^{(l)}
{p_{c,t}^{(l)}}^\top
\Bigr]
W_{\mathrm{label},t}
+
\varepsilon I .
\end{aligned}
\label{eq:metric}
\end{equation}
This metric weights projection errors by their local effect on the reference candidate-label distribution.
It provides an output-informed geometry for decomposing intermediate-layer TVs without explicitly modeling the subsequent Transformer layers.
For final-layer TVs, we use Euclidean projection, \(G_{c,t}^{(L)}=I\).
The derivation is provided in Appendix~\ref{app:hessian}.

For binary tasks, we project onto the TR and TL directions jointly using regularized least squares (Appendix~\ref{app:projection}).
For multiclass tasks, we project the mean TV onto TR and each query-wise TV onto TL.
Omitting the indices \(c,t,l\), these projections are
\begin{equation}
\begin{aligned}
v_{\mathrm{TR}}
&=
\frac{d_{\mathrm{TR}}^\top Gv_{\mathrm{TV}}}
     {d_{\mathrm{TR}}^\top Gd_{\mathrm{TR}}+\varepsilon}
d_{\mathrm{TR}},
\\
v_{\mathrm{TL},i}
&=
\frac{d_{\mathrm{TL}}(y_i)^\top Gv_{\mathrm{TV},i}}
     {d_{\mathrm{TL}}(y_i)^\top Gd_{\mathrm{TL}}(y_i)+\varepsilon}
d_{\mathrm{TL}}(y_i),
\end{aligned}
\label{eq:component_projection}
\end{equation}
where \(y_i\) is the ground-truth label of query \(i\).
We average the query-wise TL components:
\begin{equation}
v_{\mathrm{TL}}
=
\frac{1}{N}\sum_{i=1}^{N}v_{\mathrm{TL},i}.
\label{eq:multiclass_tl}
\end{equation}
For both binary and multiclass tasks, the residual is \(v_{\perp}=v_{\mathrm{TV}}-v_{\mathrm{TR}}-v_{\mathrm{TL}}\).

\subsection{TV Optimization}
\label{sec:optimization}

The decomposition above separates the original TV into functionally motivated components, but does not show whether these components provide useful control over task behavior.
We therefore search for combinations of TR and TL coefficients that improve task performance, with the residual coefficient fixed at one.

We parameterize the candidate intervention vector using the TR and TL coefficients while retaining the residual component:
\begin{equation}
u_{c,t}^{(l)}(\alpha,\beta)
=
\alpha v_{\mathrm{TR},c,t}^{(l)}
+
\beta v_{\mathrm{TL},c,t}^{(l)}
+
v_{\perp,c,t}^{(l)}.
\label{eq:candidate_tv}
\end{equation}
Retaining the residual preserves the part of the original TV not assigned to the TR and TL components and guarantees
\(u_{c,t}^{(l)}(1,1)=v_{\mathrm{TV},c,t}^{(l)}\).

To separate directional changes from magnitude changes, we normalize the candidate intervention vector before intervention.
For an optimization query \(x_j\), the vector added to the hidden state is:
\begin{equation}
\Delta h_j^{(l)}
=
\lambda
\left\lVert h_{j,\mathrm{last}}^{(l)}\right\rVert_2
\frac{
u_{c,t}^{(l)}(\alpha,\beta)
}{
\left\lVert u_{c,t}^{(l)}(\alpha,\beta)\right\rVert_2
},
\label{eq:intervention}
\end{equation}
where \(\lambda\) is a fixed global scaling coefficient.
The effective intervention magnitude is therefore normalized according to the target hidden-state norm.
We optimize only \((\alpha,\beta)\) using Gaussian-process Bayesian optimization with accuracy on \(\mathcal{D}_{\mathrm{opt}}\) as the objective, independently for each layer.


\section{Experiments}
\label{sec:experiments}
We evaluate whether the proposed decomposition supports both effective control and functional interpretation of TVs.
Our experiments examine accuracy improvements from coefficient optimization, the output-level effects of the individual components, and the effectiveness of TR/TL-based search relative to random alternatives.

\paragraph{Models and Datasets.}
To evaluate the effectiveness of our proposed method, we applied it to Llama 3.2 1B~\cite{meta2024llama32}, Llama 3.1 8B~\cite{grattafiori2024llama3}, Qwen 3.5 2B Base, Qwen 3.5 2B, and Qwen 3.5 9B~\cite{qwen2026qwen35} on four classification datasets: SST-2~\cite{socher2013recursive}, 
SUBJ~\cite{pang2004sentimental}, Financial PhraseBank (FP)~\cite{malo2014good}, and AG News~\cite{zhang2015character}.
These tasks contain 2, 2, 3, and 4 candidate labels, respectively.

\paragraph{TV Construction and Intervention.}
For demonstrations, we used four examples per label.
TVs were averaged over 256 construction queries.
During inference, we intervened on the final-token hidden state using Eq.~\ref{eq:intervention}, i.e., add the $\Delta h_j^{(l)}$ to the final-token hidden state of a specific layer $l$, with the intervention strength fixed to \(\lambda=1.0\).

\paragraph{Direction Optimization.}
We used Gaussian-process Bayesian optimization over \((\alpha,\beta)\in[0,2]^2\), with 12 initial evaluations followed by 30 optimization iterations.
Further implementation details are provided in Appendix~\ref{app:bo}.

\paragraph{Baselines.}
We compared TR/TL-based optimization that decomposes the original TV along sampled directions and optimizes their coefficients $\alpha, \beta$ while retaining the residual, with two random controls: original TV and Random.
The \textbf{original TV}, corresponding to \((\alpha,\beta)=(1,1)\), served as the primary baseline.
\textbf{Random} uses Gaussian directions, whereas Random$_\perp$ restricts them to the orthogonal complement of the original TR/TL subspace.
Both controls used the same coefficient range, BO budget, and intervention normalization as our method, with five seeds for direction sampling.
Construction details are provided in Appendix~\ref{app:random_controls}.

\paragraph{Evaluation.}
Demonstrations, TV construction queries, and evaluation queries did not overlap.
For independent evaluation, we selected the layer and coefficients maximizing optimized accuracy on the dataset used for optimization $D_{\mathrm{opt}}$ and fixed them before testing on 500 disjoint examples per task.
We evaluated the original TV at the same selected layer; details are provided in Appendix~\ref{app:independent_eval}.

\begin{figure*}[t]
    \centering
    \includegraphics[width=\textwidth]{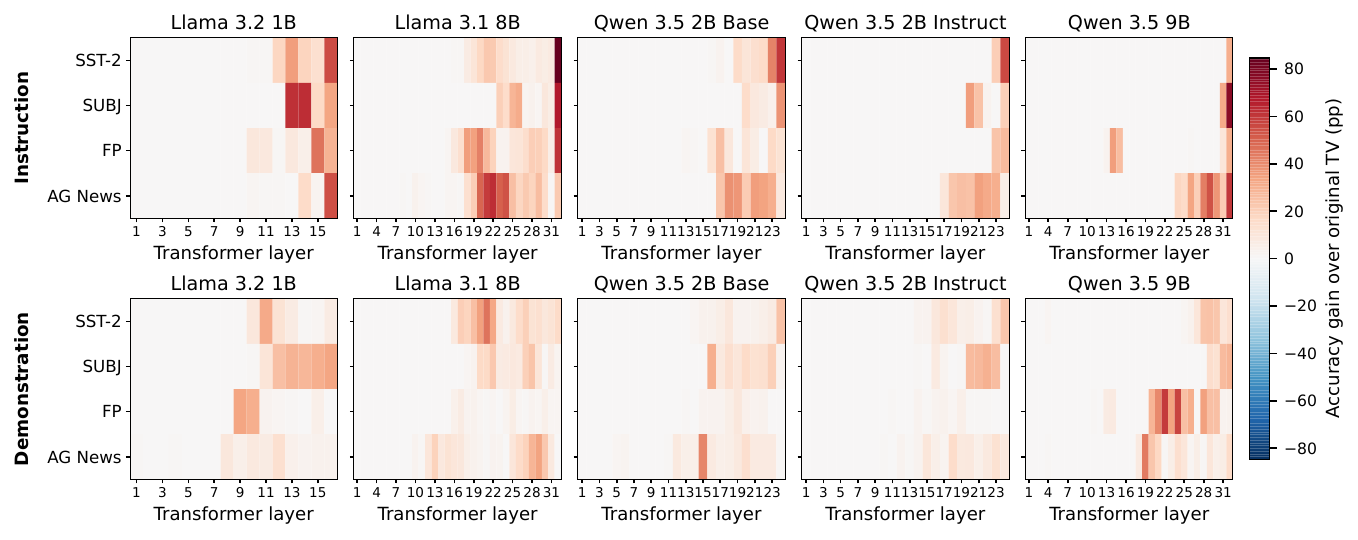}
    \vspace{-2\baselineskip}
    \caption{
    Accuracy gain over the original TV on the optimization set across Transformer layers.
    Each cell shows the change in accuracy (percentage points) after TR/TL-based optimization; red and blue indicate improvement and degradation, respectively.
    All Transformer layers are shown.
    }
    \label{fig:main_results}
    \vspace{-0.6\baselineskip}
\end{figure*}

\begin{figure*}[t!]
    \centering
    \includegraphics[width=0.96\textwidth]{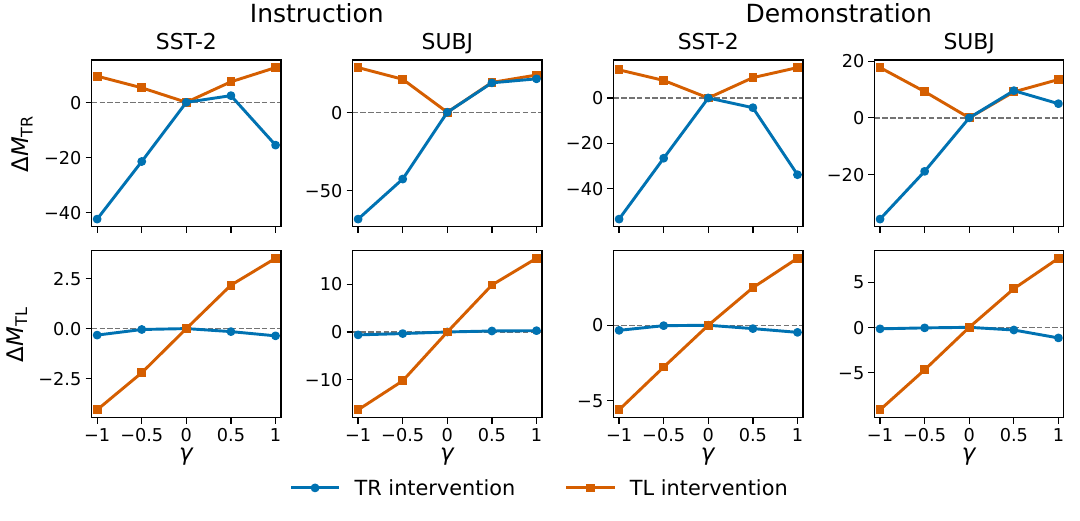}
    \vspace{-1.2\baselineskip}
    \caption{Effects of isolated TR and TL component interventions on the TR- and TL-related logit metrics.}
    \label{fig:component_effects}
\end{figure*}

\subsection{TR/TL-Based Optimization Improves TV Accuracy}

We first evaluate whether optimizing the TR and TL coefficients improves the original TV across models, tasks, and TV construction conditions.
Fig.~\ref{fig:main_results} compares the optimized and original TVs at each Transformer layer.
Improvements appear for both instruction- and demonstration-based TVs and across multiple model families and tasks, with gains ranging from modest increases to substantial improvements in some settings.

Improvements are concentrated in the middle and later layers, with little or no gain in early layers. This is aligned with prior analyses, which show weaker alignment between states and label unembedding directions in early layers~\cite{yang2025unifying}, and also suggest that even TR or TL directions are explicitly specified and could propagate as effective control components at the output layer, the absence of available task vectors at these layers prevents good initialization for the optimization.

The layer-wise results above are measured on the optimization set.
We also evaluated the selected configurations on independent examples to examine whether the improvements persist beyond the data used for optimization.
For Llama 3.2 1B and Llama 3.1 8B, optimized TVs largely outperform the original TVs at the selected layers across all four tasks and both TV construction conditions.
Full results are provided in Appendix~\ref{app:independent_eval}.


\subsection{TR/TL Decomposition Captures Distinct Functional Roles}

The preceding results establish that optimizing the component coefficients improves task performance.
We next examine whether the components affect model outputs in ways consistent with their intended roles.
The TR component is intended to increase preference for all the candidate labels over other vocabulary tokens while largely preserving their relative preference.
The TL component is intended to change the preference for the correct label over the other candidates.
We tested these expected roles through isolated component interventions on binary classification SST-2 and SUBJ.

For each task and TV construction condition, we selected the layer at which the original TV achieves the highest accuracy and intervene on the query-only hidden state using either \(v_{\mathrm{TR}}\) or \(v_{\mathrm{TL}}\) alone.
For component \(v_c\), where \(c\in\{\mathrm{TR},\mathrm{TL}\}\), we apply
\begin{equation}
h_{\gamma}
=
h
+
\gamma
\lVert h\rVert_2
\frac{v_c}{\lVert v_c\rVert_2},
\label{eq:component_intervention}
\end{equation}
with \(\gamma\in\{-1,-0.5,0,0.5,1\}\).
Let \(\mathcal{Y}\) denote the candidate-label set, \(y\) the correct label, \(\bar{y}\) the incorrect label, and \(z\) the vocabulary logits.
We then measure two complementary logit-level effects:
\begin{equation}
M_{\mathrm{TR}}(z)
=
\log\sum_{j\in\mathcal{Y}}e^{z_j}
-
\log\sum_{k\notin\mathcal{Y}}e^{z_k},
\label{eq:tr_metric}
\end{equation}
which measures how strongly the model favors the candidate labels over the rest of the vocabulary, and
\begin{equation}
M_{\mathrm{TL}}(z)
=
z_y-z_{\bar{y}},
\label{eq:tl_metric}
\end{equation}
which measures the preference for the correct label over the incorrect one.
For each metric, we report the change from the no-intervention condition, \(\Delta M=M(z_\gamma)-M(z_0)\).

Fig.~\ref{fig:component_effects} shows agreement between the intended roles and the observed responses across both tasks and TV construction conditions.
TR interventions substantially change candidate-label preference (top row, blue) while leaving label discrimination nearly unchanged (bottom row, blue).
TL interventions consistently change label discrimination with intervention strength (bottom row, orange).
These responses provide empirical support for the functional interpretation of the decomposition: the label-common and label-contrast constructions yield components with the corresponding output-level effects in the model.

The correspondence also has limits.
Candidate-label preference responds non-monotonically to TR intervention in some settings (e.g., SST-2), so increasing intervention strength does not uniformly strengthen the intended effect. This is why we use Bayesian optimization for a moderate intervention rather than simply increasing it.
TL interventions also change candidate-label preference, which is compatible with their intended role because the label-set metric aggregates the candidate logits.
Together, these results show that the decomposition captures the intended functional distinction, while its control over model outputs is not uniformly monotonic across the intervention magnitude, echoing our design of Bayesian optimization.


\begin{figure}[t]
    \centering
    \includegraphics[width=\linewidth]{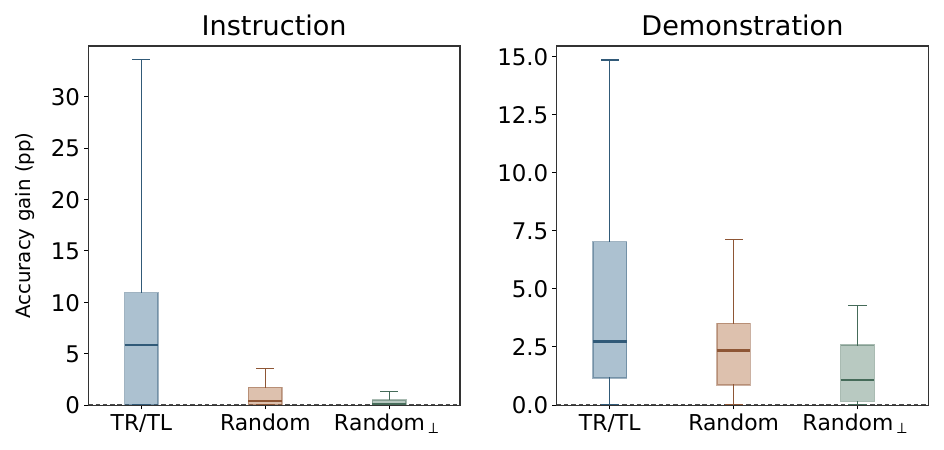}
    \vspace{-1.7\baselineskip}
    \caption{Accuracy gains over the original TV using TR/TL and random search axes.}
    \label{fig:random_comparison}
    \vspace{-0.7\baselineskip}
\end{figure}

\subsection{TR/TL Axes Provide a More Reliable Search Space Than Random Axes}

The preceding results show that the TR/TL decomposition supports performance improvement and distinct output-level effects.
As a brief ablation experiment, we next test whether its choice of directions leads to larger gains than decomposing the same original TV along random directions.
All methods retain their corresponding decomposition residual and recover the original TV at $(\alpha,\beta)=(1,1)$.
For each model-task pair and TV construction condition, we first select the layer at which the original TV achieves its highest accuracy, without using any post-optimization performance for layer selection.
At this fixed layer, we computed the accuracy gain of each method relative to the original TV.
For each random control, we averaged the gain over five independently sampled vector configurations.
Since the random directions may accidentally contain components aligned with the TR/TL subspace, to control for this possibility, Random$_\perp$ samples random directions from the orthogonal complement of the subspace spanned by the identified TR and TL directions.
All search methods are compared at the same layer selected using the original TV.

In the results shown in Fig.~\ref{fig:random_comparison}, random search axes can also improve the original TV, indicating that performance-improving directions are not unique to the TR/TL decomposition.
However, TR/TL-based search yields clearly larger gains for instruction-based TVs, while the difference is smaller for demonstration-based TVs.
Notably, for demonstration-based TVs, unconstrained random directions yield larger gains than Random$_\perp$.
This pattern is consistent with the possibility that part of the improvement from random search arises when randomly sampled directions overlap with the TR/TL subspace.
Overall, arbitrary low-dimensional search can uncover better TVs, but the TR/TL decomposition provides a more effective and functionally structured search space, particularly for instruction-based TVs.


\begin{table}[t]
\centering
\vspace{1\baselineskip}

\small
\setlength{\tabcolsep}{3pt}
\renewcommand{\arraystretch}{1.2}

\begin{tabular}{
    @{}
    p{0.24\linewidth}
    p{\dimexpr0.38\linewidth-2\tabcolsep\relax}
    p{\dimexpr0.38\linewidth-2\tabcolsep\relax}
    @{}
}
\toprule
\textbf{Component}
& \textbf{With gains}
& \textbf{Without gains}
\tabularnewline
\midrule

Random 1
& \raggedright positive, negative, neutral, \ldots
& \raggedright quo, ince, feit, keit, croft
\tabularnewline

\addlinespace[4pt]

Random 2
& \raggedright neutral, positive, negative, \ldots
& \raggedright anon, $\sigma$, paha, uese, adget
\tabularnewline

\bottomrule
\end{tabular}
\vspace{-0.7\baselineskip}
\caption{
Top-scoring tokens for random components in representative runs.
}
\label{tab:random_readout}
\end{table}

\paragraph{Qualitative Analysis: Why Can Random Directions Improve Performance?}

Random directions can also improve accuracy, although they are not constructed from candidate-label geometry.
We inspected representative improving and non-improving directions to explore their relation to the candidate labels.
We applied the curvature term of the label-based metric to each random component and then mapped the resulting vector to token scores using the full-vocabulary unembedding matrix; details are provided in Appendix~\ref{app:hessian}.
Under this readout, the improving example ranks task-related tokens, including candidate labels, among the top-scoring tokens, whereas the non-improving example shows less task-specific structure (Table~\ref{tab:random_readout}).
These examples illustrate differences under a label-based readout, reflecting the definition of TR, while the detailed mechanisms behind their accuracy gains remain unclear.


\begin{figure}[t]
    \centering
    \includegraphics[width=\linewidth]{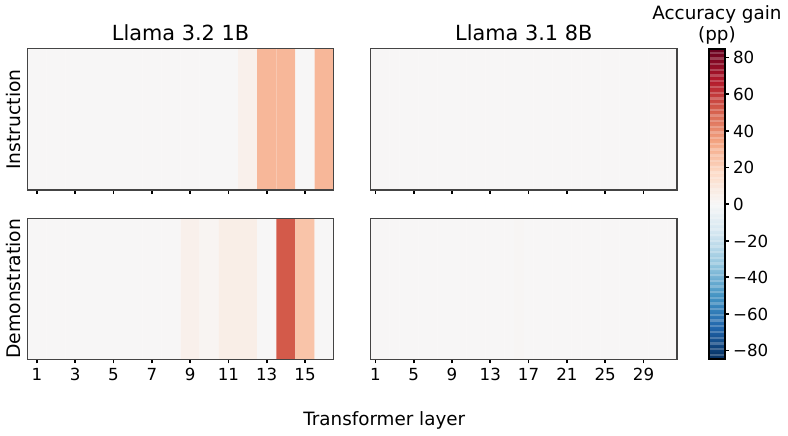}
    \vspace{-1.7\baselineskip}
    \caption{Accuracy gains over original TV on JF-ICR.}
    \label{fig:jficr_results}
    \vspace{-0.7\baselineskip}
\end{figure}

\subsection{When Does Direction Search Fail?}
\label{sec:jficr}

The preceding experiments show improvements on standard classification benchmarks.
To examine the same optimization in a more challenging setting, we evaluate Llama 3.2 1B and Llama 3.1 8B on JF-ICR~\cite{peng2026ebisu}.
This task requires recognizing implicit commitments and refusals in financial responses written in Japanese, a language not officially supported by either model.

Fig.~\ref{fig:jficr_results} shows contrasting outcomes
across the two models.
For Llama 3.2 1B, TR/TL-based optimization improves accuracy in later layers
under both construction conditions.
Llama 3.1 8B shows little or no gain across the tested layers.

These mixed outcomes highlight a limitation:
under the tested search settings, optimizing the TR and TL coefficients
does not consistently yield better intervention directions
across models in this challenging setting. Similar to early layers, the model may fail to produce a reliable task vector, leaving no good starting point for optimization.

\section{Discussion}
\label{sec:discussion}

\paragraph{Functional dependence between TR and TL.}
Fig.~\ref{fig:component_effects} shows that the extracted TR and TL components have distinct effects on model outputs.
We consider whether this functional distinction also implies independence of the underlying computations. A simple example is whether TL can emerge spontaneously when TR is blocked. Prior work provides some clues:
Filtering the output vocabulary to the candidate labels cannot produce a correct label preference (TL) if the label set preference (TR) is absent from the logits, as shown 
in an ICL experiment on SST-2 with Llama 3 8B,~\citet{cho2025revisiting} report that direct decoding is near chance
in early layers and becomes highly accurate around layer 18.
This raises the possibility that internal task recognition helps make input information available for task-specific label discrimination.
Whether TR helps establish the label discrimination associated with TL remains an open question.

\paragraph{Balancing TR and TL.}
The estimated accuracy surfaces against $(\alpha,\beta)$ in Appendix~\ref{app:coefficient_search} (Fig.~\ref{fig:tr_tl_balance}) show high-accuracy regions with relatively small TL coefficients.
We consider why reducing the contribution of a label-discriminative component may improve prediction.
If the correct label already outranks the other candidates,
further increasing its margin does not change the prediction
within the candidate set.
Instead, the remaining requirement may be to make that label
outrank non-label tokens.
Under a fixed intervention magnitude, shifting the relative
contribution towards TR may therefore be more useful than
strengthening TL further.
A small TL coefficient does not mean that label discrimination
is absent: it may already be available in the query hidden state
or retained residual.
These observations suggest that effective control depends on
the balance needed by the existing representation, rather than
on strengthening both components.

\paragraph{Conditions for additive effects.} As mentioned before, TR and TL vectors are orthogonal but sometimes dependent, i.e., combining the components linearly need not produce additive effects on model outputs.
A local additive approximation of output log odds is appropriate when higher-order terms in the response to the coefficients are small.
Such terms may matter when task recognition changes how the input is interpreted, making the TL-related effect depend on TR.
Appendix~\ref{app:conditions} shows such analysis, but its validity remains untested.


\section{Conclusion}

This paper investigates the functional structure within a single TV and its use for interpretable control.
We derive TR and TL directions from the geometry of candidate labels and decompose the original TV into TR, TL, and residual components.
Optimizing the TR/TL magnitudes improves accuracy across multiple models and datasets, with larger gains than random directions, particularly for TVs constructed from instructions.
Isolated interventions further reveal distinct effects consistent with the intended roles of TR and TL.
These results show that the internal composition of a TV provides useful information for both understanding its behavior and improving its effectiveness.
Future work should examine how TR/TL dependencies shape the mechanism of TVs.

\clearpage
\raggedbottom
\section*{Limitations}

Our decomposition operationalizes TR and TL through candidate-label geometry.
The distinct effects of isolated component interventions support the intended functional interpretation, but do not establish that these components fully capture task recognition and task learning as studied in ICL.
In particular, controlling candidate-label preference and label discrimination does not by itself demonstrate that the corresponding internal processes have been isolated.

The projection geometry for intermediate-layer TVs is based on a direct candidate-label readout rather than the actual effect of an intervention on the final output.
It does not explicitly account for the subsequent Transformer layers and therefore provides a proxy for output sensitivity.
Consequently, the functional interpretation of the resulting components at intermediate layers is supported empirically rather than guaranteed by the projection formulation.

Our experiments reveal distinct component responses and effective coefficient settings, but do not identify the internal mechanisms that produce them.
In particular, the causes of non-monotonic TR responses, TL effects on both metrics, and high accuracy with small TL coefficients remain unresolved.
The explanations proposed in the Discussion remain hypotheses, as our experiments do not distinguish metric-related effects, nonlinear downstream processing, and dependencies between TR- and TL-related computations.

Our TR/TL-based decomposition is not guaranteed to provide the optimal directions for controlling TVs.
Rather, it offers a functionally interpretable search space motivated by mechanistic studies of ICL.
Other directions may provide more effective control.

\section*{Acknowledgment}
This work was supported by the SPDR Funding (Number 202601094073), and Programs for Junior Scientists at RIKEN.

\newpage
\bibliography{custom}

@String(ICLR = {Proc. International Conference on Learning Representations (ICLR)})

@String(ICML = {Proc. International Conference on Machine Learning (ICML)})

@String(ACL = {Proc. Annual Meeting of the Association for Computational Linguistics (ACL)})

@String(ACLFindings = {Findings of the Association for Computational Linguistics (ACL Findings)})

@String(EMNLP = {Proc. Conference on Empirical Methods in Natural Language Processing (EMNLP)})

@String(EMNLPFindings = {Findings of Empirical Methods in Natural Language Processing (EMNLP Findings)})

@String(NeurIPS = {Proc. Annual Conference on Neural Information Processing Systems (NeurIPS)})

@inproceedings{hendel2023context,
    author = "Hendel, Roee  and
      Geva, Mor  and
      Globerson, Amir",
    title = {In-Context Learning Creates Task Vectors},
    booktitle = EMNLPFindings,
    year = {2023}
}

@inproceedings{todd2024function,
    author = {Eric Todd and Millicent L. Li and Arnab Sen Sharma and Aaron Mueller and Byron C. Wallace and David Bau},
    title = {Function Vectors in Large Language Models},
    booktitle = ICLR,
    year = {2024}
}

@inproceedings{Liu2024incontext,
    author = {Liu, Sheng and Ye, Haotian and Xing, Lei and Zou, James},
    title = {In-context vectors: making in context learning more effective and controllable through latent space steering},
    booktitle = ICML,
    year = {2024}
}

@inproceedings{yang2025unifying,
    author = {Yang, Haolin and Cho, Hakaze and Zhong, Yiqiao and Inoue, Naoya},
    title = {Unifying Attention Heads and Task Vectors via Hidden State Geometry in In-Context Learning},
    booktitle = NeurIPS,
    year = {2025}
}

@inproceedings{yang2026task,
    author = {Yang, Haolin and Cho, Hakaze and Ding, Kaize and Inoue, Naoya},
    title = {Task Vectors, Learned Not Extracted: Performance Gains and Mechanistic Insights},
    booktitle = ICLR,
    year = {2026}
}

@inproceedings{min2022rethinking,
    author = {Min, Sewon  and
      Lyu, Xinxi  and
      Holtzman, Ari  and
      Artetxe, Mikel  and
      Lewis, Mike  and
      Hajishirzi, Hannaneh  and
      Zettlemoyer, Luke},
    title = {Rethinking the Role of Demonstrations: What Makes In-Context Learning Work?},
    booktitle = EMNLP,
    year = {2022}
}

@misc{wei2023larger,
      title={Larger language models do in-context learning differently}, 
      author={Jerry Wei and Jason Wei and Yi Tay and Dustin Tran and Albert Webson and Yifeng Lu and Xinyun Chen and Hanxiao Liu and Da Huang and Denny Zhou and Tengyu Ma},
      year={2023},
      eprint={2303.03846},
      archivePrefix={arXiv},
      primaryClass={cs.CL},
      url={https://arxiv.org/abs/2303.03846}, 
}

@inproceedings{pan2023context,
    author = {Pan, Jane  and
      Gao, Tianyu  and
      Chen, Howard  and
      Chen, Danqi},
    title = {What In-Context Learning ``Learns'' In-Context: Disentangling Task Recognition and Task Learning},
    booktitle = ACLFindings,
    year = {2023}
}

@inproceedings{wang2025investigating,
    author = {Wang, Xiaolei and Tang, Xinyu and Li, Junyi and Zhao, Xin and Wen, Ji-Rong},
    title = {Investigating the Pre-Training Dynamics of In-Context Learning: Task Recognition vs. Task Learning},
    booktitle = ICLR,
    year = {2025}
}

@inproceedings{yang2026localizing,
    author = {Yang, Haolin and Cho, Hakaze and Inoue, Naoya},
    title = {Localizing Task Recognition and Task Learning in In-Context Learning via Attention Head Analysis},
    booktitle = ICLR,
    year = {2026}
}

@inproceedings{gao2025shaping,
    author = {Gao, Lang  and
      Geng, Jiahui  and
      Zhang, Xiangliang  and
      Nakov, Preslav  and
      Chen, Xiuying},
    title = {Shaping the Safety Boundaries: Understanding and Defending Against Jailbreaks in Large Language Models},
    booktitle = ACL,
    year = {2025}
}

@misc{siddique2026shifting,
      title={Shifting Perspectives: Steering Vectors for Robust Bias Mitigation in LLMs}, 
      author={Zara Siddique and Irtaza Khalid and Liam D. Turner and Luis Espinosa-Anke},
      year={2026},
      eprint={2503.05371},
      archivePrefix={arXiv},
      primaryClass={cs.LG},
      url={https://arxiv.org/abs/2503.05371}, 
}

@inproceedings{han2026steer2adapt,
    author = {Pengrui Han and Xueqiang Xu and Keyang Xuan and Peiyang Song and Siru Ouyang and Runchu Tian and Yuqing Jiang and Cheng Qian and Pengcheng Jiang and Jiashuo Sun and Junxia Cui and Ming Zhong and Ge Liu and Jiawei Han and Jiaxuan You},
    title = {Steer2Adapt: Dynamically Composing Steering Vectors Elicits Efficient Adaptation of {LLM}s},
    booktitle = {ICLR 2026 Workshop on Representational Alignment (Re{\textasciicircum}4-Align)},
    year = {2026}
}

@misc{agarwal2026boreft,
      title={BOReFT: Manifold Steering of Language Models for Black-box Optimization}, 
      author={Dhruv Agarwal and Rico Angell and Kavitha Srinivas and Tahira Naseem and Horst Samulowitz and Willie Neiswanger and Andrew McCallum},
      year={2026},
      eprint={2609.33722},
      archivePrefix={arXiv},
      primaryClass={cs.LG},
      url={https://arxiv.org/abs/2609.33722}, 
}

@inproceedings{cho2025revisiting,
    author = {Hakaze Cho and Mariko Kato and Yoshihiro Sakai and Naoya Inoue},
    title = {Revisiting In-context Learning Inference Circuit in Large Language Models},
    booktitle = ICLR,
    year = {2025}
}

@misc{meta2024llama32,
    author  = {{Meta AI}},
    title   = {{Llama 3.2} Model Card},
    year    = {2024},
    url     = {https://github.com/meta-llama/llama-models/blob/main/models/llama3_2/MODEL_CARD.md}
}

@misc{grattafiori2024llama3,
    title   = {The Llama 3 Herd of Models},
    author  = {Llama Team, AI @ Meta},
    year    = {2024},
    eprint  = {2407.21783},
    archivePrefix   = {arXiv},
    primaryClass    = {cs.AI},
    url     = {https://arxiv.org/abs/2407.21783}
}

@misc{qwen2026qwen35,
  title  = {{Qwen3.5}: Towards Native Multimodal Agents},
  author = {{Qwen Team}},
  month  = feb,
  year   = {2026},
  url    = {https://qwen.ai/blog?id=qwen3.5}
}

@inproceedings{socher2013recursive,
    author = {Socher, Richard  and
      Perelygin, Alex  and
      Wu, Jean  and
      Chuang, Jason  and
      Manning, Christopher D.  and
      Ng, Andrew  and
      Potts, Christopher},
    title = {Recursive Deep Models for Semantic Compositionality Over a Sentiment Treebank},
    booktitle = EMNLP,
    year = {2013}
}

@inproceedings{pang2004sentimental,
    author = {Pang, Bo  and
      Lee, Lillian},
    title = {A Sentimental Education: Sentiment Analysis Using Subjectivity Summarization Based on Minimum Cuts},
    booktitle = ACL,
    year = {2004}
}

@article{malo2014good,
  title   = {Good Debt or Bad Debt: Detecting Semantic Orientations in Economic Texts},
  author  = {Malo, Pekka and Sinha, Ankur and Korhonen, Pekka and Wallenius, Jyrki and Takala, Pyry},
  journal = {Journal of the Association for Information Science and Technology},
  volume  = {65},
  number  = {4},
  pages   = {782--796},
  year    = {2014},
  doi     = {10.1002/asi.23062}
}

@inproceedings{zhang2015character,
  title     = {Character-Level Convolutional Networks for Text Classification},
  author    = {Zhang, Xiang and Zhao, Junbo and LeCun, Yann},
  booktitle = NeurIPS,
  pages     = {649--657},
  year      = {2015}
}

@misc{peng2026ebisu,
  title   = {Ebisu: Benchmarking Large Language Models in Japanese Finance},
  author  = {Peng, Xueqing and Xiang, Ruoyu and Zhang, Fan and Song, Mingzi
             and Jiang, Mingyang and Wang, Yan and Qian, Lingfei and Hara, Taiki
             and Guo, Yuqing and Huang, Jimin and Tsujii, Junichi
             and Ananiadou, Sophia},
  year={2026},
  eprint={2602.01479},
  archivePrefix={arXiv},
  primaryClass={cs.CL},
  url={https://arxiv.org/abs/2602.01479}
}

\appendix


\section{Hessian-based Label Metric}
\label{app:hessian}
The projection metric should weight hidden-state differences according to their local effect on the reference candidate-label distribution.
We derive this metric for intermediate-layer TVs from the Hessian of the label-space log-partition function.

Given a TV \(v_{\mathrm{TV},c,t}^{(l)}\), the induced candidate-label logits are

\begin{equation}
z_{c,t}^{(l)}
=
W_{\mathrm{label},t}
v_{\mathrm{TV},c,t}^{(l)}.
\end{equation}

The corresponding label distribution is

\begin{equation}
p_{c,t}^{(l)}
=
\operatorname{softmax}
(z_{c,t}^{(l)}).
\end{equation}

For the categorical softmax function, the Hessian of the log-partition function with respect to the label logits is

\begin{equation}
H_{c,t}^{(l)}
=
\operatorname{diag}
(p_{c,t}^{(l)})
-
p_{c,t}^{(l)}
{p_{c,t}^{(l)}}^\top .
\end{equation}

The hidden-state metric is obtained by pulling this label-space curvature back through the label unembedding matrix:

\begin{equation}
G_{c,t}^{(l)}
=
W_{\mathrm{label},t}^{\top}
H_{c,t}^{(l)}
W_{\mathrm{label},t}
+
\varepsilon I .
\end{equation}

This metric emphasizes hidden-state directions that produce larger changes in the candidate-label logit distribution.
It is used as a label-logit-sensitive geometry for decomposition rather than an exact prompt-specific output metric.

For the qualitative analysis, we apply the label-derived curvature term to each random component before mapping it to vocabulary scores.
For a random component \(r\), we compute
\begin{equation}
s(r)
=
W_{\mathrm{vocab}}
W_{\mathrm{label},t}^{\top}
H_{c,t}^{(l)}
W_{\mathrm{label},t}
r,
\label{eq:random_readout}
\end{equation}
where \(W_{\mathrm{vocab}}\) is the full-vocabulary unembedding matrix.
We report the tokens with the highest scores in \(s(r)\).


\section{Projection Formulation}
\label{app:projection}

The main text defines the decomposition directions and their projection metric.
Here we derive the projection coefficients used to extract the TR and TL components and define the remaining residual.
Let

\begin{equation}
D=
\begin{bmatrix}
d_1 & d_2
\end{bmatrix}
\end{equation}

denote the matrix containing the decomposition axes.
Given a TV \(v\), the projection coefficients are obtained by solving the regularized least-squares problem:

\begin{equation}
\min_a
(v-Da)^\top
G
(v-Da)
+
\varepsilon\|a\|_2^2 .
\end{equation}

The closed-form solution is

\begin{equation}
a
=
(D^\top GD+\varepsilon I)^{-1}
D^\top Gv .
\end{equation}

For binary classification,
\(D=[d_{\mathrm{TR}},d_{\mathrm{TL}}]\)
and the resulting components are

\begin{equation}
v_{\mathrm{TR}}
=
a_1d_{\mathrm{TR}},
\qquad
v_{\mathrm{TL}}
=
a_2d_{\mathrm{TL}}.
\end{equation}

The residual component is defined as

\begin{equation}
v_\perp
=
v_{\mathrm{TV}}
-
v_{\mathrm{TR}}
-
v_{\mathrm{TL}}.
\end{equation}

For multiclass classification, the TL component is obtained by averaging query-wise projections onto the corresponding gold-vs-rest directions.


\section{Bayesian Optimization Details}
\label{app:bo}

\paragraph{Gaussian-Process Surrogate.}
We used scikit-learn's GaussianProcessRegressor with a constant kernel multiplied by an ARD Matérn-\(5/2\) kernel, plus a white-noise kernel.
The length scales are initialized to \((0.5,0.5)\) with bounds \([10^{-2},10]\), the constant multiplier to \(1\) with bounds \([10^{-3},10^3]\), and the noise variance to \(10^{-5}\) with bounds \([10^{-8},10^{-1}]\).
These hyperparameters are optimized by maximizing the log marginal likelihood using L-BFGS-B with three additional restarts.
We standardized the objective values during GP fitting and use a fixed diagonal regularization of \(10^{-8}\).

\paragraph{Acquisition Function.}
We used Expected Improvement (EI) with an exploration parameter of \(\xi=0.01\).
At each iteration, we refit the GP and evaluate EI on a fixed \(41\times41\) grid over \([0,2]^2\).
The unevaluated point with the highest EI is selected for the next evaluation.

\paragraph{Initialization and Optimization Budget.}
Each run starts with the same 12 fixed initial coefficient pairs, including $(\alpha,\beta)=(1,1)$.
This pair recovers the original TV for both the proposed method and the random controls.
The initial points are
\begin{equation}
\begin{aligned}
\{&(1,1),(0,1),(1,0),(0,0),\\
  &(2,1),(1,2),(2,2),(0.5,0.5),\\
  &(0.5,1),(1,0.5),(1.5,1),(1,1.5)\}.
\end{aligned}
\end{equation}
We performed 30 additional BO iterations, resulting in 42 evaluations per run.
The final solution is the evaluated point with the highest accuracy.

\section{Random Control Construction}
\label{app:random_controls}

Let $d$ be the hidden-state dimension and let
\[
S=\operatorname{span}\{v_{\mathrm{TR}},v_{\mathrm{TL}}\}
\]
denote the original TR/TL subspace at the selected layer.
For Random, two decomposition directions $q_1,q_2\in\mathbb{R}^{d}$ are sampled from a Gaussian distribution.
For Random$_\perp$, these directions are restricted to $S^\perp$.
For each control, we decompose the original TV along the sampled directions into components $r_1,r_2$ and an orthogonal residual:
\[
r_\perp=v_{\mathrm{TV}}-r_1-r_2,
\qquad
r_\perp\in\operatorname{span}\{q_1,q_2\}^{\perp}.
\]
The resulting intervention vector is
\begin{equation}
u_{\mathrm{rand}}(\alpha,\beta)
=\alpha r_1+\beta r_2+r_\perp.
\end{equation}
Thus, $u_{\mathrm{rand}}(1,1)=v_{\mathrm{TV}}$.
The decomposition components and residual remain fixed throughout each BO run.
We optimize $(\alpha,\beta)\in[0,2]^2$ and apply the same intervention normalization as in Eq.~\ref{eq:intervention}.
The five seeds vary the sampled decomposition directions.


\section{Independent Evaluation}
\label{app:independent_eval}

We tested whether the gains obtained during optimization persist on examples not used for BO or configuration selection.
For this purpose, we evaluated the optimized TVs on an independent set $D_{\mathrm{test}}$ containing 500 examples per task.
For each model, task, and TV construction condition, we select the layer and component coefficients with the highest optimized accuracy on $D_{\mathrm{opt}}$ and fix them before evaluation on $D_{\mathrm{test}}$, which is disjoint from $D_{\mathrm{opt}}$ and is not used for BO or configuration selection.
We compared the original and optimized TVs at the same selected layer.
As shown in Table~\ref{tab:independent_eval}, all 16 configurations retain positive accuracy gains on $D_{\mathrm{test}}$, with gain magnitudes generally comparable to those on $D_{\mathrm{opt}}$.
These results indicate that the benefits of the selected component coefficients extend to examples not used during optimization.

\begin{table*}[t]
    \centering
    \small
    \setlength{\tabcolsep}{4pt}

    \textbf{(a) Llama-3.2-1B}\par\smallskip
    \begin{tabular}{@{}llccccc@{}}
        \toprule
        \multirow{2}{*}[-0.6ex]{Task}
        & \multirow{2}{*}[-0.6ex]{TV}
        & \multirow{2}{*}[-0.6ex]{Layer}
        & \multicolumn{2}{c}{Optimization set}
        & \multicolumn{2}{c}{Independent evaluation set} \\
        \cmidrule(lr){4-5}
        \cmidrule(lr){6-7}
        & & & Original & Optimized & Original & Optimized \\
        \midrule
        \multirow{2}{*}{SST-2}
        & Instruction & 16
        & 28.91 & \accgain{84.38}{55.47}
        & 11.2 & \accgain{79.6}{68.4} \\
        & Demonstration & 13
        & 76.17 & \accgain{82.81}{6.64}
        & 80.4 & \accgain{82.8}{2.4} \\

        \multirow{2}{*}{SUBJ}
        & Instruction & 16
        & 40.23 & \accgain{73.83}{33.60}
        & 48.8 & \accgain{67.8}{19.0} \\
        & Demonstration & 16
        & 40.23 & \accgain{73.83}{33.60}
        & 48.8 & \accgain{67.8}{19.0} \\

        \multirow{2}{*}{FP}
        & Instruction & 15
        & 9.38 & \accgain{55.08}{45.70}
        & 13.0 & \accgain{56.8}{43.8} \\
        & Demonstration & 15
        & 56.64 & \accgain{61.72}{5.08}
        & 60.4 & \accgain{63.4}{3.0} \\

        \multirow{2}{*}{AG News}
        & Instruction & 16
        & 0.00 & \accgain{54.69}{54.69}
        & 0.0 & \accgain{50.6}{50.6} \\
        & Demonstration & 11
        & 60.55 & \accgain{67.58}{7.03}
        & 59.0 & \accgain{63.2}{4.2} \\
        \bottomrule
    \end{tabular}

    \par\medskip

    \textbf{(b) Llama-3.1-8B}\par\smallskip
    \begin{tabular}{@{}llccccc@{}}
        \toprule
        \multirow{2}{*}[-0.6ex]{Task}
        & \multirow{2}{*}[-0.6ex]{TV}
        & \multirow{2}{*}[-0.6ex]{Layer}
        & \multicolumn{2}{c}{Optimization set}
        & \multicolumn{2}{c}{Independent evaluation set} \\
        \cmidrule(lr){4-5}
        \cmidrule(lr){6-7}
        & & & Original & Optimized & Original & Optimized \\
        \midrule
        \multirow{2}{*}{SST-2}
        & Instruction & 32
        & 0.00 & \accgain{84.77}{84.77}
        & 0.4 & \accgain{84.2}{83.8} \\
        & Demonstration & 32
        & 68.36 & \accgain{84.38}{16.02}
        & 69.8 & \accgain{85.0}{15.2} \\

        \multirow{2}{*}{SUBJ}
        & Instruction & 31
        & 65.23 & \accgain{71.48}{6.25}
        & 55.2 & \accgain{61.6}{6.4} \\
        & Demonstration & 25
        & 64.45 & \accgain{73.44}{8.99}
        & 58.4 & \accgain{64.4}{6.0} \\

        \multirow{2}{*}{FP}
        & Instruction & 29
        & 41.80 & \accgain{62.11}{20.31}
        & 45.2 & \accgain{63.0}{17.8} \\
        & Demonstration & 25
        & 57.42 & \accgain{63.67}{6.25}
        & 61.0 & \accgain{62.8}{1.8} \\

        \multirow{2}{*}{AG News}
        & Instruction & 24
        & 16.02 & \accgain{70.31}{54.29}
        & 14.8 & \accgain{65.8}{51.0} \\
        & Demonstration & 16
        & 59.38 & \accgain{69.53}{10.15}
        & 57.6 & \accgain{67.6}{10.0} \\
        \bottomrule
    \end{tabular}

    \caption{
        Accuracy (\%) on the optimization and independent evaluation
        sets. Original and optimized TVs use the same selected layer.
        Green arrows indicate gains over the original TV in percentage
        points within each set.
    }
    \label{tab:independent_eval}
\end{table*}


\section{Conditions and Scope of Linear Control}
\label{app:conditions}

Our results show that optimizing the TR and TL coefficients can improve an original TV while preserving a functional interpretation of the components.
A central question is when this linear combination can also be understood as a combination of their effects on model outputs.
This requires considering both how the model responds to intervention and how the task depends on recognizing its purpose and interpreting its inputs.

\paragraph{Conditions for additive effects.}
Let \(p_{\alpha,\beta}(y\mid x)\) denote the output distribution under the intervention in Eq.~\ref{eq:intervention}, including its normalization.
For an output \(y\) and a reference output \(y_0\), define the log odds
\(L_{x,y}(\alpha,\beta)=\log[p_{\alpha,\beta}(y\mid x)/p_{\alpha,\beta}(y_0\mid x)]\).
These measure the relative support for the two outputs.
Assuming a smooth response, a first-order Taylor approximation around the original TV gives
\begin{equation}
\begin{aligned}
L_{x,y}(\alpha,\beta)
\approx {}&
L_{x,y}(1,1)\\
&+(\alpha-1)a_{x,y}
+(\beta-1)b_{x,y},
\end{aligned}
\label{eq:discussion_additivity}
\end{equation}
where \(a_{x,y}\) and \(b_{x,y}\) are the derivatives with respect to the TR and TL coefficients at \((1,1)\).
This approximation describes each component as contributing a fixed change to the log odds.
It is accurate when higher-order terms are small over the coefficient changes being considered.
In particular, a small mixed term means that changing one component does not substantially alter the contribution of the other.
In probability space, additive effects on log odds correspond to multiplying the relative contributions of the components and normalizing the resulting distribution.

\paragraph{Dependence between task recognition and interpretation.}
A linguistic interpretation of this condition is that recognizing the task leaves the relevant interpretation of the input largely stable.
For example, a model may already understand the sentiment of a sentence, while needing an adjustment to how strongly it favors the candidate labels or reflects that judgment in its label preference.
Such a correction can plausibly be achieved by changing the strength of existing contributions.
A stronger dependence arises when recognizing the task changes which aspects of the input matter.
A politely worded response, for example, may convey positive sentiment while declining a request.
If changing TR alters which interpretation the model uses, the contribution associated with TL may also change.
A fixed additive description would then be insufficient over that range of interventions.
This dependence need not remain at every layer: the model may already have resolved the relevant interpretation before the intervention is applied.

\paragraph{Relation to the empirical results.}
The accuracy gains show that adjusting the existing components can correct some errors across multiple settings.
The distinct effects of isolated component interventions further support their functional interpretation.
The effect of TL on both \(M_{\mathrm{TL}}\) and \(M_{\mathrm{TR}}\) is compatible with additive contributions: two components can affect the same output metric without changing each other's contribution.
However, neither this observation nor improved accuracy establishes that interactions are small.
The approximation above therefore provides a conditional explanation of how the components can combine, rather than a property established throughout the search range.

\paragraph{Scope of the decomposition.}
The two coefficients control adjustments to information already encoded in the model and the original TV, including the retained residual.
Their usefulness depends on whether the required correction can be expressed by strengthening or weakening the selected contributions.
Corrections that require other directions, or changes in how the contributions depend on each other, can exceed this description.
The contrasting outcomes on JF-ICR illustrate that the usefulness of this restricted adjustment depends on the model even for the same task, although they do not identify the cause of the difference.
The relevant scope of linear control is therefore conditional on the learned representation: complex linguistic dependencies may already be represented, leaving an adjustment that a small number of interpretable components can support.


\section{TR/TL Coefficient Search}
\label{app:coefficient_search}

\begin{figure}[H]
    \centering
    \begin{minipage}[t]{0.88\linewidth}
        \centering
        \includegraphics[width=\linewidth]{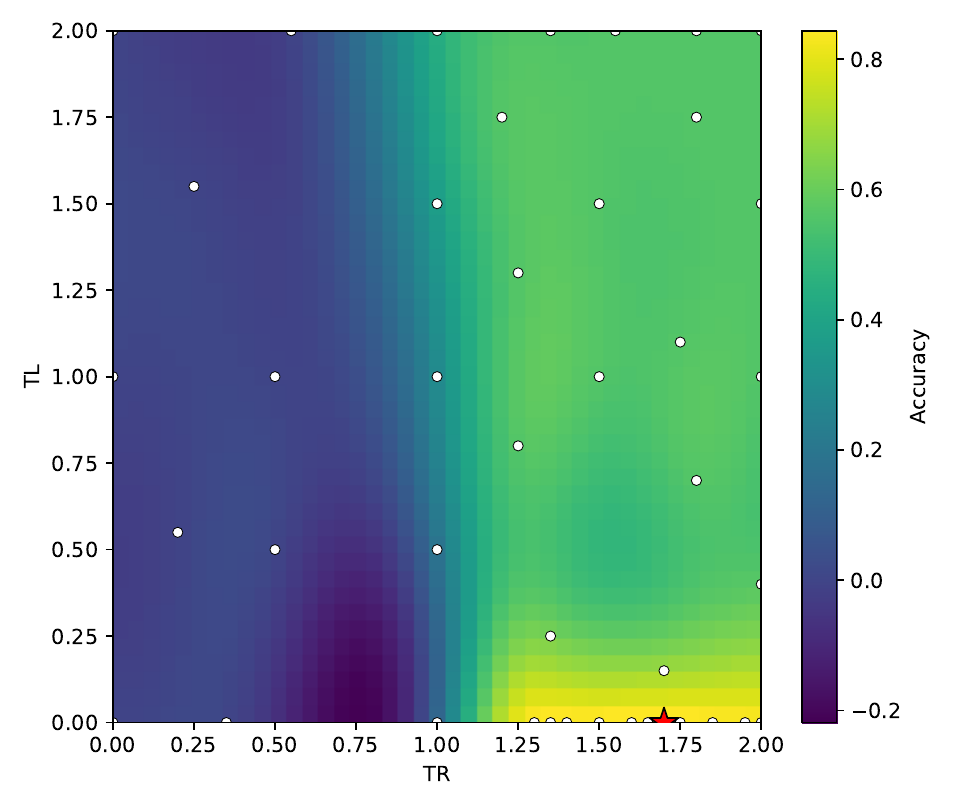}
        \par\smallskip
        {\small (a) SST-2}
    \end{minipage}
    \medskip
    \begin{minipage}{0.85\linewidth}
        \centering
        \includegraphics[width=\linewidth]{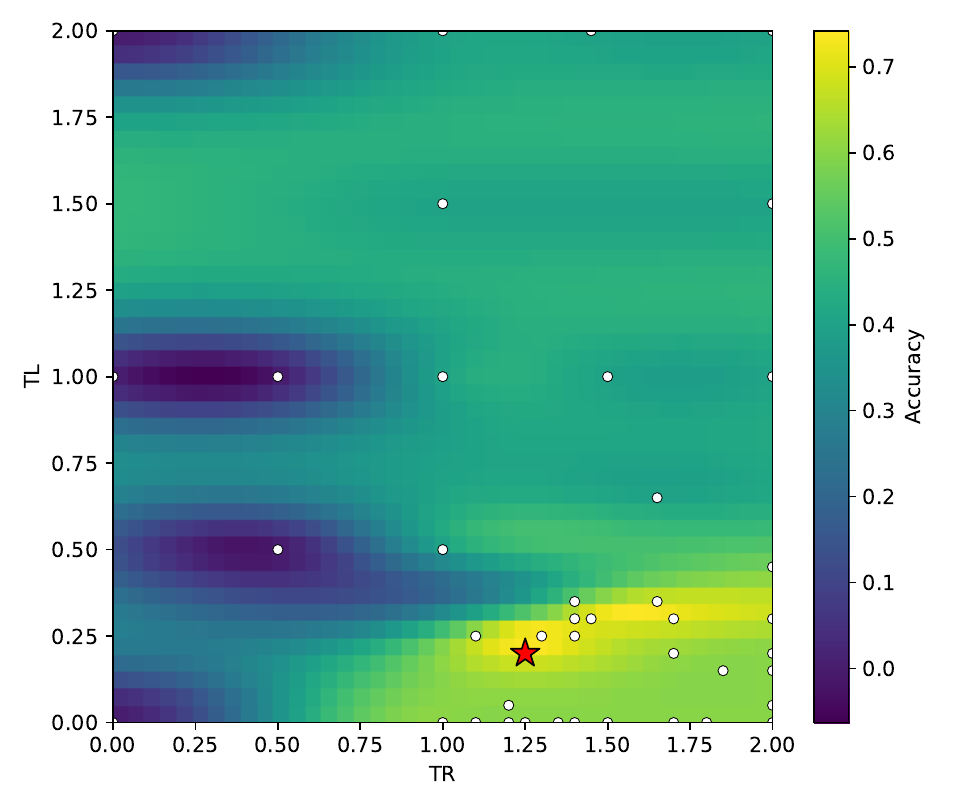}
        \par\smallskip
        {\small (b) SUBJ}
    \end{minipage}
    \caption{
        GP-predicted accuracy across TR and TL coefficients.
        White circles indicate evaluated coefficient settings;
        red stars indicate the selected optimal settings.
    }
    \label{fig:tr_tl_balance}
\end{figure}

Fig.~\ref{fig:tr_tl_balance} shows the GP-estimated accuracy surfaces over the TR and TL coefficients for instruction-based TVs on SST-2 and SUBJ with Llama-3.2-1B.
Both surfaces contain high-accuracy regions with relatively large TR coefficients and small TL coefficients, although the shapes of these regions differ between tasks.
These examples motivate the discussion of TR/TL balance in Section~\ref{sec:discussion}, while the GP estimates alone do not establish why these coefficient settings are effective.

\end{document}